\documentclass[conference]{IEEEtran}
\IEEEoverridecommandlockouts
\usepackage[l2tabu,orthodox]{nag}

\usepackage[
    backend=bibtex,
    style=ieee,
    sorting=none,
    natbib=true,
    doi=false,
    isbn=false,
    url=false,
    eprint=false,
    maxcitenames=1,
    mincitenames=1
]{biblatex}

\usepackage[pdftex,colorlinks]{hyperref}
\usepackage[printonlyused]{acronym}

\usepackage{siunitx}
\usepackage[all]{nowidow}

\usepackage[dvipsnames]{xcolor}

\usepackage{lipsum}

\usepackage{xspace} %smart handling of space in commands

\usepackage{makecell}

\usepackage[pdftex]{graphicx}

\usepackage{epstopdf}

\usepackage{import}

\usepackage{booktabs}

\usepackage{tabularx}
\usepackage{multirow, multicol}

\usepackage{amssymb,amsfonts,amsmath,amscd}

\usepackage{bm}

\newcommand{\norm}[1]{\left\Vert#1\right\Vert}

\newcommand{\bbm}{\begin{bmatrix}}
\newcommand{\ebm}{\end{bmatrix}}

\newcommand{\ignore}[1]{}

\newcommand{\bma}[1]{\left[\begin{array}{#1}}
\newcommand{\ema}{\end{array}\right]}

\DeclareMathAlphabet{\mbf}{OT1}{ptm}{b}{n}

\newcommand{\mbfhat}[1]{{\hat{\mbf{#1}}}}

\newcommand{\mbfcheck}[1]{{\check{\mbf{#1}}}}

\def\fdotb{{\raisebox{-0.6ex}{ \kern0.2ex\raisebox{0.8ex}{\tiny $\hspace*{-1ex}\circ$}}}}
\def\fddotb{{\raisebox{-0.6ex}{ \kern0.2ex\raisebox{0.8ex}{\tiny $\hspace*{-1ex}\circ\circ$}}}}

\newcommand{\trans}{{\ensuremath{\mathsf{T}}}} % transpose
\newcommand{\utimes}{ {\raisebox{-0.6ex}{ \kern-1.0ex\raisebox{0.6ex}{ \small $\mathsf{v}$}}} } % 
\newcommand{\beq}{\begin{equation}}
\newcommand{\eeq}{\end{equation}}
\newcommand{\bdis}{\begin{displaymath}}
\newcommand{\edis}{\end{displaymath}}
\newcommand{\beqarray}{\begin{eqnarray}}
\newcommand{\eeqarray}{\end{eqnarray}}
\newcommand{\beqarraynn}{\begin{eqnarray*}}
\newcommand{\eeqarraynn}{\end{eqnarray*}}

\DeclareMathAlphabet{\mbf}{OT1}{ptm}{b}{n}

\usepackage{tikz}
\usetikzlibrary{arrows}
\usetikzlibrary{arrows.meta}
\usetikzlibrary{shapes}

\usepackage{algorithmic}
\usepackage{textcomp}
\def\BibTeX{{\rm B\kern-.05em{\sc i\kern-.025em b}\kern-.08em
    T\kern-.1667em\lower.7ex\hbox{E}\kern-.125emX}}
    
\begin{document}

\title{Pointing the Way: \\Refining Radar-Lidar Localization \\Using Learned ICP Weights\\
\thanks{This work was supported by the Province of Ontario QEII-GSST and OGS scholarships and the NSERC CGS D scholarship. Resources used in preparing this research were provided, in part, by the Province of Ontario and the Government of Canada through their support of the Vector Institute.}
}

\author{\IEEEauthorblockN{Daniil Lisus, Johann Laconte, Keenan Burnett, Ziyu Zhang, and Timothy D. Barfoot}
\IEEEauthorblockA{\textit{Robotics Institute}\\
\textit{University of Toronto}\\
Toronto, Canada\\
Email: [FIRSTNAME].[LASTNAME]@robotics.utias.utoronto.ca}
}

\maketitle

\begin{abstract}
This paper presents a novel deep-learning-based approach to improve localizing radar measurements against lidar maps.
This radar-lidar localization leverages the benefits of both sensors; radar is resilient against adverse weather, while lidar produces high-quality maps in clear conditions.
However, owing in part to the unique artefacts present in radar measurements, radar-lidar localization has struggled to achieve comparable performance to lidar-lidar systems, preventing it from being viable for autonomous driving.
This work builds on ICP-based radar-lidar localization by including a learned preprocessing step that weights radar points based on high-level scan information.
To train the weight-generating network, we present a novel, stand-alone, open-source differentiable ICP library.
The learned weights facilitate ICP by filtering out harmful radar points related to artefacts, noise and even vehicles on the road.
Combining an analytical approach with a learned weight reduces overall localization errors and improves convergence in radar-lidar ICP results run on real-world autonomous driving data.
Our code base is publicly available to facilitate reproducibility and extensions\footnote[1]{\,Paper code is available at: https://github.com/utiasASRL/mm\_masking.}.
\end{abstract}


\section{INTRODUCTION}
    \label{sec:introduction}
    Recent autonomous vehicle progress has been facilitated in part by the collection of detailed maps of commonly visited locations.
When preconstructed maps are available, the primary vehicle navigation task is simplified to finding the vehicle pose within the map, commonly referred to as \textit{localization}.
An accurate and reliable localization estimate is critical for safe autonomous operation.

\begin{figure}[t]
    \centering
    \begin{tikzpicture}
        \node (figure) at (0.0,0) {\includegraphics[width=\linewidth]{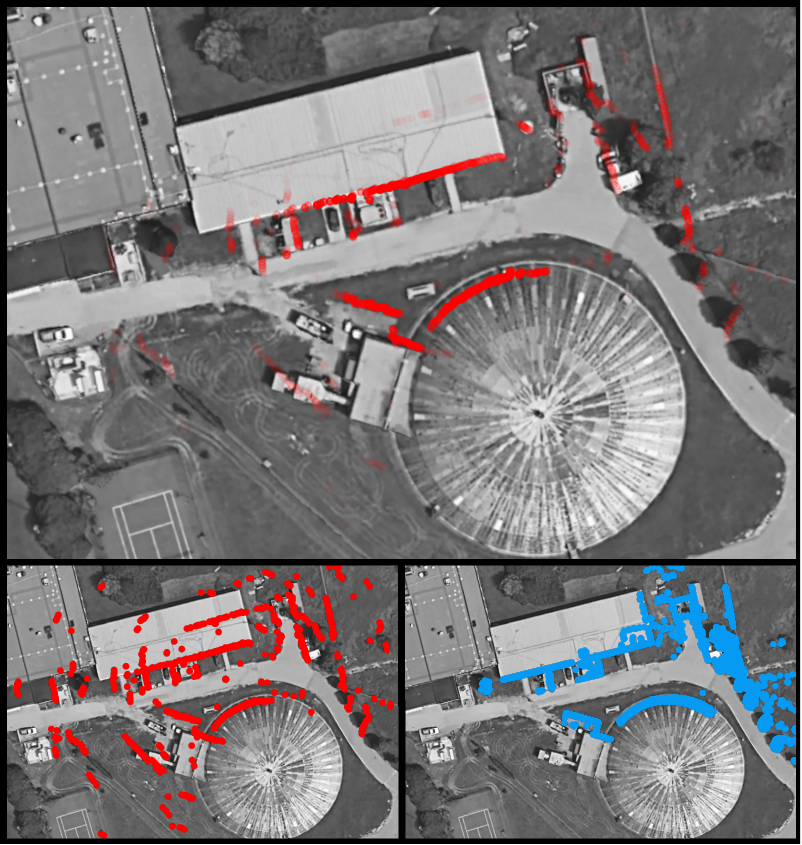}};
        \node[draw=black, rounded corners, fill=white, inner sep=2pt] (lidar) at (-2.4,4.0) {{\textbf{Weighted Radar Scan}}};
        \node[draw=black, rounded corners, fill=white, inner sep=2pt] (full_scan) at (-2.8,-4.1) {{\textbf{Full Radar Scan}}};
        \node[draw=black, rounded corners, fill=white, inner sep=2pt] (weighted_scan) at (1.1,-4.1) {{\textbf{Lidar Map}}};

        % Highlight key areas
        % Echo removed
        \node[draw=black, rounded corners, fill=white, inner sep=2pt] (echo) at (-1.4,2.9) {{\scriptsize \textcolor{black}{\textbf{Echo removed}}}};
        \node[draw=orange, rotate=-76, line width=0.3mm, minimum width=0.42cm, minimum height=1.3cm, shape=ellipse] (echo_pts) at (0.4,3.45) {};
        \draw[-{Triangle[length=1.4mm,width=1.8mm]}, draw=orange, line width=0.3mm] (echo) -- (echo_pts);
        % Noise removed
        \node[draw=black, rounded corners, fill=white, inner sep=2pt] (noise) at (2.0,4.2) {{\scriptsize \textcolor{black}{\textbf{Noise removed}}}};
        \node[draw=orange, line width=0.3mm, minimum width=0.42cm, minimum height=1.0cm, shape=circle] (noise_pts) at (3.6,3.9) {};
        \draw[-{Triangle[length=1.4mm,width=1.8mm]}, draw=orange, line width=0.3mm] (noise) -- (noise_pts);
        % Walls highlighted
        \node[draw=black, rounded corners, fill=white, inner sep=2pt] (walls) at (1.5,0.0) {{\scriptsize \textcolor{black}{\textbf{Walls highlighted}}}};
        \draw[-{Triangle[length=1.4mm,width=1.8mm]}, draw=orange, line width=0.3mm] (walls) -- (0.7,1.3);
        \draw[-{Triangle[length=1.4mm,width=1.8mm]}, draw=orange, line width=0.3mm] (walls) -- (0.25,0.75);
        
    \end{tikzpicture}
    
    \caption{A satellite image of the environment, overlaid with the radar pointcloud weighted using our method (red, top), full extracted radar pointcloud (red, bottom), and lidar map (blue). The weight associated with each point in the weighted pointcloud corresponds to the intensity in colour. Our method picks out most of the important geometric features while ignoring noisy points and radar artefacts, thus facilitating the task of ICP. }
    \label{fig:intro}
\end{figure}
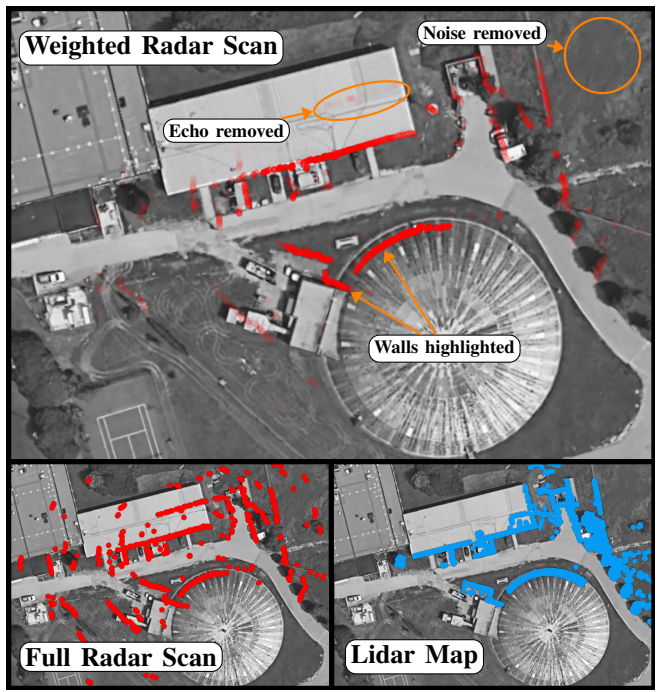

The current state of the art for the task of localization is matching live lidar scans to preconstructed lidar maps \cite{survey_lidar_loc}.
Lidar sensors are capable of capturing high-definition maps, which can then be used to achieve accurate localization.
However, lidar measurements can be affected or completely blocked by precipitation and other challenging environmental conditions \cite{are_we_ready_for, yoneda2019automated, Courcelle_Baril_Pomerleau_Laconte_2022, Park_Kim_Kim_2019}.
Radar has been considered as an alternative for localization in all-weather conditions.
Radar, owing to its longer wavelength, is much less susceptible to small particles in the air, and is thus largely unaffected by rain, snow, dust, or fog \cite{weather_sensor_impact}.
However, the longer wavelengths and lower resolution inherently lead to sparser maps as compared to those constructed using lidar.
Additionally, radar measurements suffer from artefacts such as speckle (noisy measurements), saturation (falsely strong return signal), and ghosting (multiple echoes of the same point or object) \cite{radar_nav_book}.
These artefacts are challenging to model \cite{Weston_Jones_Posner_2020}.
As a result, radar-radar localization accuracy is still worse than that of lidar-lidar \cite{are_we_ready_for}, making it not viable for tasks that require centimeter-level accuracy, such as autonomous driving.

Owing to the unique advantages and disadvantages of lidar and radar, there has been interest in multi-modal radar-lidar localization.
By localizing using radar scans against already available high-quality lidar maps, the goal is to achieve near state-of-the-art localization performance, while being resilient to all weather conditions.
However, past work in radar-lidar localization has not been able to achieve the accuracy required for autonomous driving \cite{are_we_ready_for, RaLL, Park_Kim_Kim_2019, RoLM}.
Our previous work, \cite{are_we_ready_for}, was able to achieve state-of-the-art radar-lidar localization by using the iterative closest point (ICP) algorithm to match radar and lidar pointclouds, but still fell short of desired localization accuracy \cite{icp_ref}. 

Our hypothesis is that unique radar artefacts and noise properties significantly degrade the quality of radar-lidar localization.
To combat this, we refine ICP-based radar-lidar localization using a learned weight mask to identify only the most useful radar points for the task of radar-lidar localization. This learned filtering approach, visualized in Figure~\ref{fig:intro}, is the primary contribution of this paper.
Additionally, we provide an open-source, stand-alone, differentiable ICP library for future approaches that wish to incorporate learned elements in an ICP pipeline.

The rest of this paper is structured as follows. Section \ref{sec:related_work} presents relevant prior work. Section \ref{sec:methodology} details our approach. Finally, Section \ref{sec:experiments} provides experimental results, with concluding remarks presented in Section \ref{sec:conclusion}.

\section{RELATED WORK}
    \label{sec:related_work}
    Radar has seen a resurgence within robotics in recent years \cite{radar_survey, radar_survey_2, venon2022millimeter}. 
The ability of modern-day radar systems to produce reasonable long-term odometry results is shown in \cite{Cen_Newman_2018}, where an ICP-style pointcloud alignment is used to compute the relative motion between consecutive timestamps.
An alternative approach is presented in \cite{masking_by_moving}, where a network is trained to mask raw radar scans before using them for correlative scan matching of consecutive frames.
It is shown that such a learned approach is capable of outperforming heuristic filtering methods.
A network that directly extracts features, descriptors, and feature scores from raw radar images and then uses them to match radar data from consecutive timesteps is presented in \cite{under_the_radar}.
A self-supervised feature learning approach, which does not require groundtruth data, is shown in \cite{hero_paper}.
Although feature-extraction methods are typically more interpretable and robust to poor initial guesses, they frequently require large networks with higher computational cost.
Additionally, feature matching across sensing modalities, such as between radar and lidar, poses an extra challenge.
Finally, most radar algorithms have focused on odometry or simultaneous localization and mapping (SLAM), e.g., \cite{Jose_Adams_2004, Rouveure_Monod_Faure_2009, Checchin_Gerossier_Blanc_Chapuis_Trassoudaine_2010, Schuster_Keller_Rapp_Haueis_Curio_2016}, meaning that investigations into the performance of radar in localization against a previously known map is understudied.
An extended Kalman filter (EKF) ICP-based approach for radar-radar localization is presented in \cite{Ward_Folkesson_2016}. However, their localization errors are on the order of meters, far above autonomous driving requirements.
In our recent ICP-based approach \cite{are_we_ready_for}, localization is performed using batch optimization with a modified Constant False Alarm Rate (CFAR) \cite{cfar} point extractor called Bounded False Alarm Rate (BFAR) \cite{bfar}. This work shows that radar localization against an existing radar map can approach the level needed for autonomous driving \cite{reid2019localization}, with translation errors of $6-10 \, \si{\cm}$.

Radar has also been used to localize against lidar maps.
One of the first algorithms for radar-lidar localization was proposed for indoor disaster environments where lidar is blocked by smoke \cite{Park_Kim_Kim_2019}.
This algorithm implements a recursive undistortion approach to utilize shape matching between live radar scans and an existing lidar map.
This recursive implementation is shown to overcome some aspects of radar noise and artefacts to achieve usable localization results in situations where lidar measurements completely fail.
A learning-based approach was taken in  \cite{radar_on_lidar}, where a generative adversarial network (GAN) was used to style-transfer a live radar scan to lidar data.
Instead of doing a cross-sensor style transfer, Yin \textit{et al.} \cite{RaLL} use a set of networks to transfer both radar and lidar scans to a common learned embedding in which they then perform cross-correlation matching.
This match is then treated as a measurement in a Kalman-filter framework.
Their algorithm is set up to be fully differentiable, meaning that the networks can be trained to produce an embedding that directly minimizes the final localization output from the Kalman filter.
Another embedding-style approach is presented in \cite{RoLM}, where it is hypothesized that although extracted radar and lidar points do not necessarily have one-to-one correspondences, their groupings are related.
Thus, radar points are binned into discrete Cartesian and polar groupings and an initial alignment between radar and lidar groupings is done via a density metric.
The estimate is then further refined using ICP.
A learned radar-lidar place recognition algorithm is presented in \cite{Yin_Xu_Wang_Xiong_2021}, but they do not evaluate localization accuracy.
Although promising, none of these papers are able to achieve centimeter-level localization accuracy, a requirement for safe autonomous vehicle operations.
However, our previous work was able to achieve an accuracy on the order of tens of centimeters with a nonlearned approach \cite{are_we_ready_for}.
This algorithm first converts the radar and lidar measurements to pointclouds using BFAR, and then applies ICP to match the pointclouds.

\begin{figure*}[t]
    \centering
    \includegraphics[width=\linewidth]{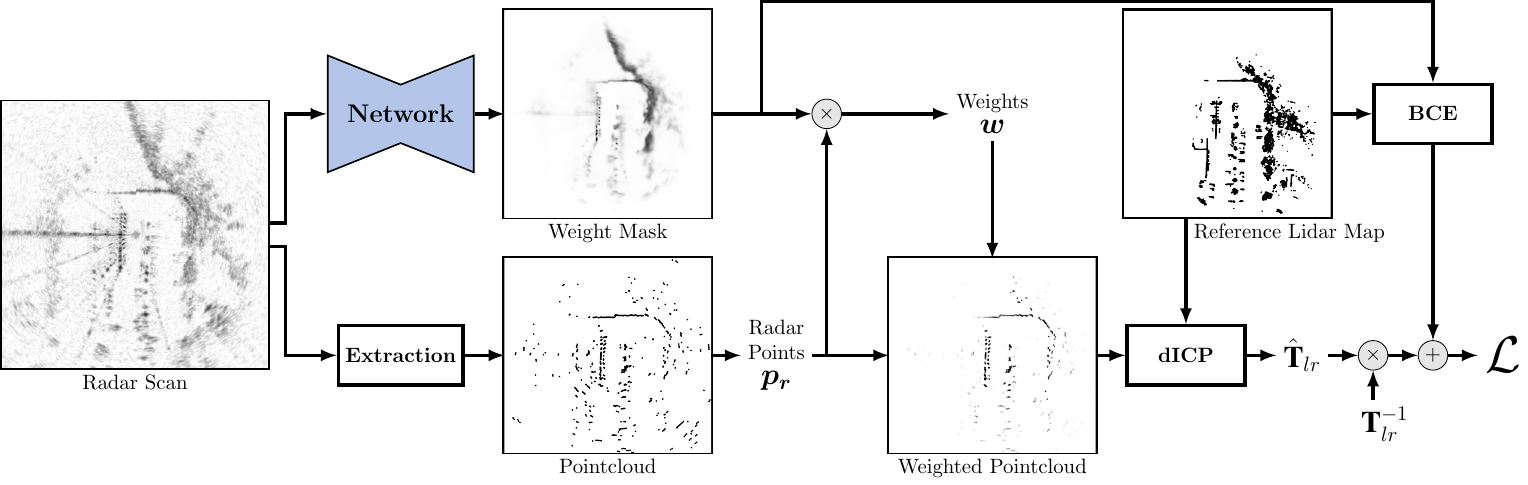}
    \caption{The pipeline used to train the weight mask network. A radar scan is fed to a U-Net network that produces a weight mask. In parallel, the scan is used to extract a pointcloud using the BFAR detector. The pointcloud then indexes weights from the weight mask. The points and corresponding weights are used in a weighted differentiable ICP algorithm (dICP) to localize against the reference map pointcloud. The training loss is formed from the error in the dICP result and a binary cross-entropy (BCE) term computed between the weight mask and a supervisory map mask. The network is trained by backpropagating the loss through dICP and the weight extraction.}
    \label{fig:pipeline}
\end{figure*}

In our previous work, we have found that CFAR-based point detectors work well for generating radar pointclouds. These can subsequently be matched to prebuilt lidar maps using ICP to yield state-of-the-art radar-lidar performance. 
This approach makes use of the full-resolution radar scan without needing to downsample, as in \cite{hero_paper}, and does not require any discretization of the pose estimate, as in \cite{RaLL, masking_by_moving}.
However, CFAR-like detectors only use local information to extract points, meaning that a significant amount of noise gets extracted.
Thus, our approach is to start with the radar-lidar localization pipeline presented in \cite{are_we_ready_for} and, building on \cite{masking_by_moving}, incorporate a learned weight mask into the ICP algorithm to identify only the most useful radar points for the task of radar-lidar localization.
Using a network allows us to learn the high-level spatial contextual cues in order to softly reject points that may correspond to noise, while still retaining the high-resolution, discretization-free point localization from CFAR and ICP.
\section{METHODOLOGY}
    \label{sec:methodology}
    \subsection{Weighted Point-to-point ICP}
This section presents the specific ICP implementation that our pipeline aims to improve.
The goal of ICP is to align an $n$-point, $D$-dimensional source pointcloud $\mbf{S} \in \mathbb{R}^{n \times D}$ to an $m$-point, $D$-dimensional target pointcloud $\mbf{P} \in \mathbb{R}^{m \times D}$. ICP solves this alignment problem in an iterative fashion to find the transformation $\mbf{T}_{ts} \in SE(D)$, where $SE(D)$ is the special Euclidean group of dimension $D$ \cite{barfoot2017state}. In our case, the source pointcloud is the radar scan, the target pointcloud is the lidar map, and $D = 2$. 

First, for each point $\mbf{s}_i$ in $\mbf{S}$, we find the nearest neighbour $\mbf{q}_i$ from $\mbf{P}$. The definition of `nearest' is a design choice, and we only present the so-called point-to-point definition as it is the one used in this paper. However, our method could be easily be applied to other common definitions, for example point-to-plane, without substantial changes. In the case of point-to-point, we compute $\mbf{q}_i$ as
\begin{align}\label{eq:nearest_neighbour}
    \mbf{q}_i = \mathrm{argmin}_{\mbf{q} \in \mbf{P}} \left(\norm{\mbfcheck{T}_{ts} \mbf{s}_i - \mbf{q}}_2\right) \in \mathbb{R}^{D},
\end{align}
where we use the latest estimate $\mbfcheck{T}_{ts}$ and abuse notation to have $\mbfcheck{T}_{ts} \mbf{s}_i \in \mathbb{R}^{D}$. An error is then formed between $\mbf{s}_i$ and $\mbf{q}_i$ as
\begin{align}
    \mbf{e}_i = \mbfcheck{T}_{ts}\mbf{s}_i - \mbf{q}_i \in \mathbb{R}^{D}.
\end{align}
A robust-cost function $\rho(\cdot) \colon \mathbb{R}^{D} \rightarrow \mathbb{R}^{D}$ can be used to down-weight large outlier errors by re-scaling the error as $\rho(\mbf{e}_i)$~\cite{robust_func}.
A trimmed ICP modification, which we use here, can be included into $\rho(\cdot)$ to zero-out error terms that have a magnitude above some threshold.
If prior knowledge about the reliability of specific points as compared to others exists, a weight $\mbf{w}_p \in \mathbb{R}^{n \cdot D}$ is included in the optimization problem.
The final objective function is then
\begin{align}\label{eq:ls_problem}
    J(\mbf{T}_{ts}) = \frac{1}{2} \mbf{e}^\trans \mbf{W} \mbf{e} \in \mathbb{R},
\end{align}
where $\mbf{e} \in \mathbb{R}^{n \cdot D}$ is a stacked vector of all $\mbf{e}_i$, $\rho(\mbf{e})$ is applied error-wise for an element-wise multiplication $\odot$ in $\mbf{W} = \mathrm{diag}(\rho(\mbf{e}) \odot \mbf{w}_p) \in \mathbb{R}^{n \cdot D \times n \cdot D}$, and which can be minimized using least-squares to find an optimal $\mbfhat{T}_{ts}$. This paper aims to find an optimal $\mbf{w}_p$.

\subsection{Architecture}
Our overall pipeline architecture is presented in Figure \ref{fig:pipeline}.
The input to the pipeline is a single radar intensity scan.
A Cartesian representation of the scan, normalized to have a maximum pixel value of one, is used as input to the network, whereas an unnormalized polar representation is fed to the point-extraction module.
Inspired by \cite{masking_by_moving}, we use a U-Net network~\cite{unet} to learn the weight mask for each radar input.
A U-Net lends itself well to our application, as we are, in essence, trying to segment the radar scan into `good' and `bad' areas for pointcloud extraction.
The encoder increases the number of channels from one, the measured intensity, to 8, 16, 32, 64, 128, and 256.
Each channel increase is composed of convolution blocks that include a 2D convolutional layer (kernel size 3, padding 1), a ReLU activation, another 2D convolution, a dropout layer with probability of 0.05, and finally a max-pooling operation (kernel size 2, stride 2).
Using dropout helps to prevent over-fitting and encourages the network to learn higher-level, general information instead of relying on specific pixels \cite{dropout}.
The decoder then decreases the channel number from 256 down to 8 using the same steps as the encoder.
For each decrease, the input is up-sampled, passed through the same convolution block as described for the encoder, concatenated with the corresponding skip connection from the encoder section, and passed through a convolution block again.
Finally, the output is passed through one last 2D convolutional layer to decrease the 8 channels back down to one, and then through a sigmoid activation.
The resulting mask is then normalized to have a maximum value of one.
The network is implemented in PyTorch \cite{pytorch}.

In parallel to the weight-mask generation, a pointcloud is extracted from the unnormalized, polar radar intensity scan using the BFAR detector.
This pointcloud is then used to index the weight mask using bilinear interpolation between pixels.
The extracted weight values, alongside the radar pointcloud, are then fed into the weighted dICP algorithm to be aligned with a given lidar map pointcloud.

Note, that as radar is a 2D sensor, the overall pipeline runs in 2D.
This means that the dICP algorithm estimates a transform from the radar scan to the lidar map, $\mbfhat{T}_{lr} \in SE(2)$, which is contrasted with the groundtruth pose, $\mbf{T}_{lr} \in SE(2)$, to form the ICP error vector $\mbf{e}_\mathrm{ICP}$ according to
\begin{align}
    \mbf{e}_\mathrm{ICP} = \bbm {e}_x & {e}_y & {e}_\phi \ebm^\trans = \log\left(\mbfhat{T}_{lr}^{}\mbf{T}_{lr}^{-1}\right)^\vee,
\end{align}
where $\log(\cdot)$ is the logarithmic operator that converts $SE(2)$ elements to their Lie algebra representation and $(\cdot)^\vee$ is the operator that converts the Lie algebra representation to $\mathbb{R}^3$~\cite{barfoot2017state}. Additionally, ${e}_x$, ${e}_y$, and ${e}_\phi$ are the $x$ (longitudinal), $y$ (lateral), and $\phi$ (heading) error components, respectively.

The loss is then formed based on two signals.
The first, $\mathcal{L}_\mathrm{ICP}$, is based on $\mbf{e}_\mathrm{ICP}$, and is formed as
\begin{align}
\mathcal{L}_\mathrm{ICP} = \mbf{e}_\mathrm{ICP}^\trans \bbm \alpha & 0 & 0 \\ 0 & \alpha & 0 \\ 0 & 0 & \beta \ebm \mbf{e}_\mathrm{ICP},
\end{align}
where $\alpha,\beta\in\mathbb{R}_{\geq 0}$ control the relative impact of each translation and rotation term.

The second loss signal, $\mathcal{L}_\mathrm{BCE}$, is a binary cross-entropy loss computed between the generated weight mask and a supervisory map mask discussed in Section \ref{sec:sup_map_mask}. The final loss term is then of the form
\begin{align}\label{eq:loss}
    \mathcal{L} = \mathcal{L}_\mathrm{ICP} + \gamma \mathcal{L}_\mathrm{BCE},
\end{align} 
where $\gamma\in\mathbb{R}_{\geq 0}$ controls the relative impact of $\mathcal{L}_\mathrm{BCE}$ on $\mathcal{L}$.

After training is complete, weight masks can be generated from single radar scans during inference time.
The subsequent weighted pointcloud can be fed to any ICP algorithm.

\subsection{Differentiable ICP}
A key part of the pipeline is the differentiable ICP (dICP) algorithm, which is needed to backpropagate directly from the ICP error. This algorithm is made publicly available\footnote[2]{\,dICP code is available at: https://github.com/utiasASRL/dICP} and adds to the recent corpus of open-source, differentiable versions of classic robotic algorithms \cite{pypose, grad_slam, binbin_paper}. Indeed, all three of these libraries include a differentiable ICP implementation. However, they are all restrictive in that they only implement either point-to-point or point-to-plane cost terms, do not make use of any robust cost function or only allow for one type, do not accept weighted pointclouds, and, for \cite{grad_slam, binbin_paper}, are embedded within a SLAM pipeline, without a clear way to run ICP in a stand-alone fashion.

Our dICP implementation, based in PyTorch, is entirely stand-alone and light-weight, making it simple to plug into any project requiring a differentiable ICP component.
We provide support for both point-to-point and point-to-plane ICP. We also include options for either a Huber or Cauchy loss and make it easy to incorporate additional losses.
It is also possible to specify a trimming parameter to run trimmed ICP, and to provide weights for the source pointcloud in order to run weighted ICP.
Depending on the problem at hand, it is possible to toggle dICP to run in 2D or 3D.
Finally, the differentiability of the algorithm can be toggled on and off, in case it is desired to run approximation-free ICP during inference.
The main components of dICP are described in the following subsections.

\subsubsection{Nearest Neighbour}
The first important step of each ICP iteration is to find the nearest point in the target pointcloud from each source pointcloud point.
Here, `nearest' refers to the smallest $L_2$ norm distance.
Although computing the $L_2$ distance between each source point and every map point is differentiable, picking out the minimum index from this result is not.
Of note is the fact that the act of indexing is differentiable if the index remains constant, as this amounts to a constant matrix multiplication.
However, since the nearest neighbour for each point changes as ICP iterates, the index is not constant.
One popular approximation for the $\mathrm{argmax}$ operation, or $\mathrm{argmin}$ in our case as in \eqref{eq:nearest_neighbour}, is the Gumbel-Softmax \cite{gumbel_softmax}.
This approximation turns the distances into probabilities and then computes a softmax across these probabilities to find the `soft' nearest neighbour.
However, as this `soft' neighbour is computed based on an average across all probabilities, it ceases to correspond to a true, physical map point, potentially limiting how finely the source pointcloud may be aligned to the target. An additional downside is that Gumbel-Softmax greatly increases the computation graph, slowing down training and requiring more GPU memory.

As such, we also provide and use an alternative setting to treat the nearest neighbour as a locally constant operation, and still making use of $\mathrm{argmin}$.
Consider one source point $\mbf{s}$ that is well aligned with the target point $\mbf{p}_1$, to which it actually corresponds.
Shifting $\mbf{s}$ in the neighbourhood of $\mbf{p}_1$ will continue to yield $\mbf{p}_1$ as the nearest neighbour, meaning the operation can be viewed as constant.
Moreover, gradient descent will be effective in minimizing the distance between $\mbf{s}$ and $\mbf{p}_1$.
Now, consider that $\mbf{s}$ is instead in the vicinity of a different target point $\mbf{p}_2$.
Alone, $\mbf{s}$ will locally converge to $\mbf{p}_2$ if gradient descent is run.
However, the assumption of ICP is that, on average, the source pointcloud will be drawn towards the target pointcloud through local convergences to neighbours. This will eventually drive $\mbf{s}$ into the neighbourhood of $\mbf{p}_1$, where the nearest-neighbour operation becomes truly constant and a fine alignment becomes possible.
Although this approach prevents the network from learning through the nearest-neighbour operation, which can be detrimental if the initial transformation is very off, we find that it works very efficiently in our training where the initial transformation is set to the groundtruth (see Section \ref{sec:TVT}).

% This approach performs well in the context of this paper.
% However, studying the tradeoff between using an approach like Gumbel-Softmax and treating the nearest-neighbour operation as constant is left for future work.

\subsubsection{Error Computation} Using the nearest neighbours, an error term based on either point-to-point or point-to-plane is computed.
This operation is fully differentiable.

\subsubsection{Trim Distance} If a trim distance is specified, the difference between the norm of the distance between each point and its nearest neighbour is computed.
This difference is then fed into a $\mathrm{tanh}$ function transformed to scale inputs between zero and one.
The result is a weight smoothly transitioning from one for points that are within the trim distance to zero for points that are outside.
A threshold is used when differentiability is turned off.

\subsubsection{Robust Loss}
Next, a robust-loss function can be specified.
A pseudo-Huber loss is used for the Huber loss during differentiable computations \cite{hartley2003multiple}, while the Cauchy loss remains the same as it is already differentiable.

\subsubsection{Update Step}
Finally, a weighted least-squares problem is formed as in \eqref{eq:ls_problem}.
Gradient descent is used to update the estimated transform.
If 2D ICP is required, extra dimensions in the update step are set to zero.
Gradient descent is, by definition, differentiable and thus poses no issues.

\subsubsection{Iteration}
Finally, the above steps are repeated until convergence or a maximum number of iterations is reached. Each iteration simply adds additional links in the PyTorch computational graph, meaning that any number of iterations are possible. However, in practice, the maximum number of iterations may be restricted by available memory.

\subsection{Supervisory Map Mask}
\label{sec:sup_map_mask}
In order to better constrain the learning and help initial convergence, a supervisory signal for the weight mask based on the reference map is added to the loss term.
To generate this reference, each map point is first transformed into the radar sensor frame using the groundtruth. All points in the range of the radar scan are then projected to a Cartesian grid, with pixel values nearest to the projected points being set to one, while all others are set to zero. This `map mask' is then compared to the weight mask produced by the network through a binary cross-entropy loss, leading to the loss term $\mathcal{L}_{\mathrm{BCE}}$ in \eqref{eq:loss}. We find this single supervisory signal sufficient to stabilize the learning process and ensure a more robust weight mask without needing to add additional terms such as a point-to-point correspondence loss. 

\section{EXPERIMENTS}
    \label{sec:experiments}
    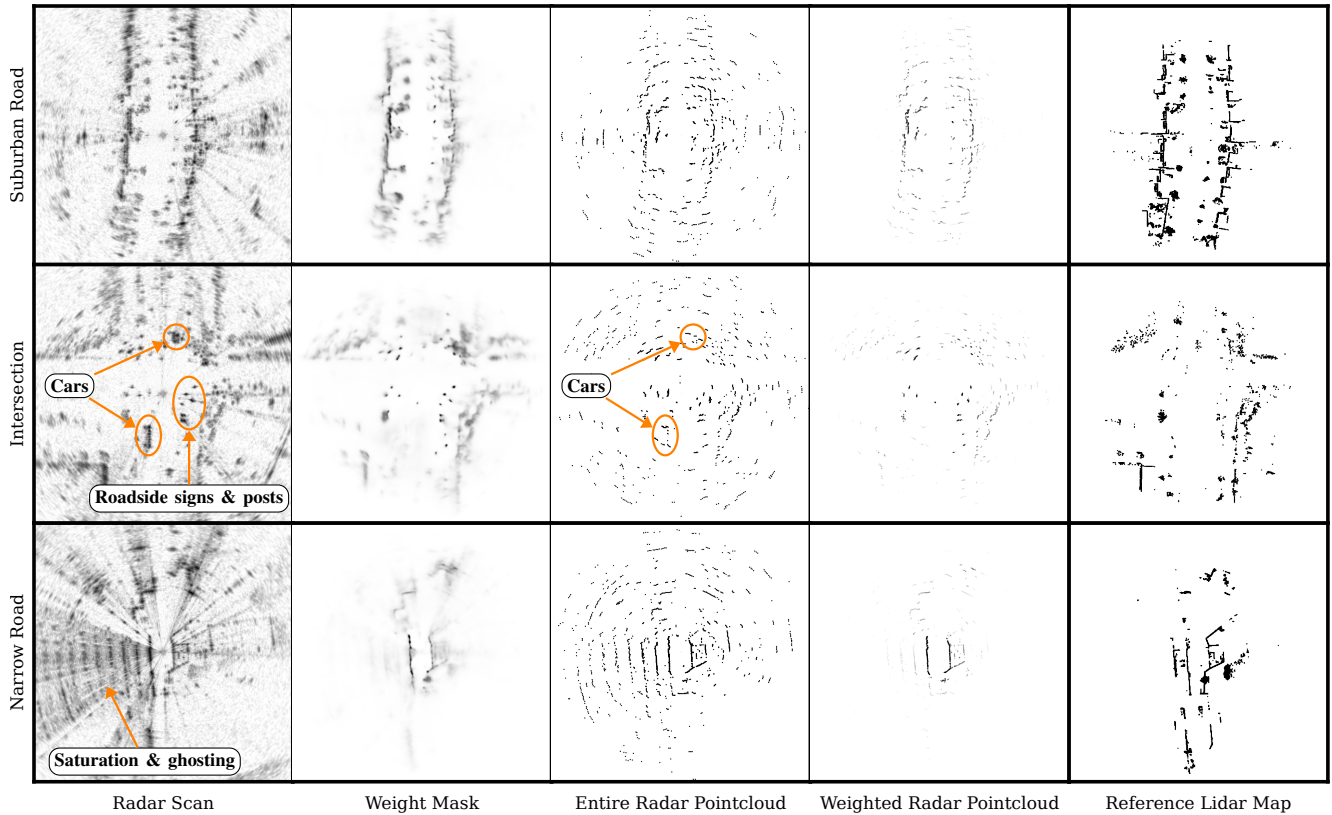
\begin{figure*}[t!]
    \centering
    \begin{tikzpicture}
        % Intersection
        % Road signs and etc. in radar scan
        \node (figure) at (0.0,0) {\includegraphics[width=\linewidth]{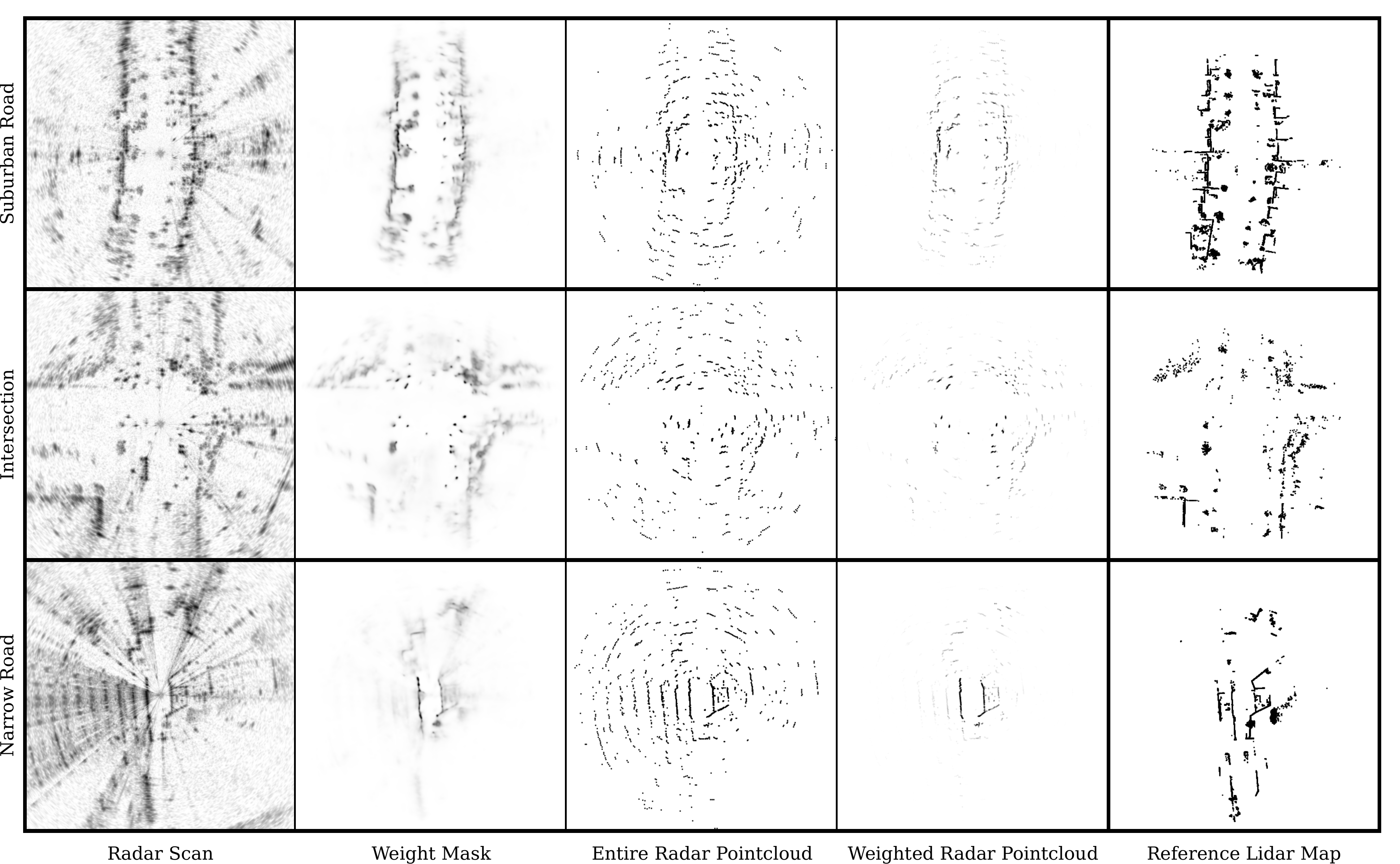}};
        \node[draw=black, rounded corners, fill=white, inner sep=2pt] (roadside) at (-6.66,-1.3) {{\scriptsize \textcolor{black}{\textbf{Roadside signs \& posts}}}};
        \node[draw=orange, line width=0.3mm, minimum width=0.42cm, minimum height=0.7cm, shape=ellipse] (signs) at (-6.66,-0.03) {};
        \draw[-{Triangle[length=1.4mm,width=1.8mm]}, orange, line width=0.3mm] (roadside) -- (signs);
        % Cars in radar scan
        \node[draw=black, rounded corners, fill=white, inner sep=2pt] (cars) at (-8.25,0.2) {{\scriptsize \textcolor{black}{\textbf{Cars}}}};
        \node[draw=orange, line width=0.3mm, minimum width=0.1cm, minimum height=0.53cm, shape=ellipse] (truck) at (-7.20, -0.46) {};
        \draw[-{Triangle[length=1.4mm,width=1.8mm]}, orange, line width=0.3mm] (cars) -- (truck);
        \node[draw=orange, line width=0.3mm, minimum width=0.15cm, minimum height=0.15cm, shape=ellipse] (car) at (-6.83,0.85) {};
        \draw[-{Triangle[length=1.4mm,width=1.8mm]}, orange, line width=0.3mm] (cars) -- (car);
        % Cars in entire radar pointcloud
        \node[draw=black, rounded corners, fill=white, inner sep=2pt] (cars_2) at (-1.23,0.2) {{\scriptsize \textcolor{black}{\textbf{Cars}}}};
        \node[draw=orange, line width=0.3mm, minimum width=0.15cm, minimum height=0.53cm, shape=ellipse] (truck_2) at (-0.18, -0.46) {};
        \draw[-{Triangle[length=1.4mm,width=1.8mm]}, orange, line width=0.3mm] (cars_2) -- (truck_2);
        \node[draw=orange, line width=0.3mm, minimum width=0.15cm, minimum height=0.15cm, shape=ellipse] (car_2) at (0.19,0.85) {};
        \draw[-{Triangle[length=1.4mm,width=1.8mm]}, orange, line width=0.3mm] (cars_2) -- (car_2);
        
        % Narrow road
        % Artefacts in radar scan
        \node[draw=black, rounded corners, fill=white, inner sep=2pt] (artefacts) at (-7.1,-4.8) {{\scriptsize \textcolor{black}{\textbf{Saturation \& ghosting}}}};
        \draw[-{Triangle[length=1.4mm,width=1.8mm]}, orange, line width=0.3mm] (artefacts) -- (-7.6, -3.8);
    \end{tikzpicture}
    \caption{Three examples of the output from the pipeline in different scenarios. The mask highlights nearby structured areas when they exist (suburban road), focuses on reliable roadside signs and posts (intersection), all while ignoring cars (intersetion) and unique radar artefacts (narrow road).}
    \label{fig:example_masks}
\end{figure*}

\subsection{Dataset}
We train, validate, and test the architecture using the Boreas dataset \cite{boreas}, in which the same path is repeatedly traversed over the course of a year by a car equipped with a radar and lidar.
To generate samples for this paper, we use the topometric localization pipeline from \cite{are_we_ready_for}.
This pipeline is based on the VT\&R framework \cite{VTR}, where a sequence is used to construct a series of connected submaps against which repeated traversals are localized.
We first construct lidar submaps based on one traversal, and then collect `live' radar scans during subsequent traversals.
Whenever the pipeline ran ICP for radar-lidar localization, the radar scan, radar pointcloud, and map pointcloud were saved.

\subsection{Training, Validation, and Testing}
\label{sec:TVT}
To mirror \cite{are_we_ready_for}, we used point-to-point ICP with a trim distance of $5.0 \, \si{\meter}$ and a Cauchy robust loss with a parameter of $1.0$. We trained using the Adam \cite{adam} optimizer with a learning rate of $1e^{-4}$ and a batch size of 5. Approximately $50,000$ samples from 12 trajectories were used for training.
The loss-weighting parameters were set to $\alpha=\beta=\gamma=1$.
Since we wish to improve on localization results that already achieve a low error, we initialized all initial guesses to the groundtruth.
More precisely, we used the groundtruth to transform the map pointcloud into the radar frame, and then used identity as the initial guess for all samples.
This additionally helped to keep the number of dICP iterations at 10 during training, greatly reducing the memory requirements and improving epoch speed.
To encourage the network to learn orientation-agnostic masks, the radar scan and aligned pointclouds were randomly rotated between $0\si{\degree}$ to $360\si{\degree}$.
This random augmentation happened every time a sample was loaded, meaning that the same localization pair was rotated differently every epoch.
Finally, we noted that the small batch size led to training instability on account of poorly converging ICP samples dominating the gradient in some batches.
To combat this, we only back-propagated through `good' samples: those that converged to a step size below $0.01$ and had an ICP error norm below $0.4$. 

Validation used $2,000$ samples taken from 3 additional trajectories.
The differentiability in dICP was turned off and the maximum number of iterations was incresed to 50 to better match practical ICP algorithms. The initial guess was still initialized to the groundtruth.

Finally, the best weights were used to run tests on 6 additional full trajectories, testing approximately $27,000$ frames.
dICP was run in nondifferentiable mode and given 50 iterations to converge.
To study the effects of the learned weight mask in practical localization applications, the lateral, longitudinal, and heading components of the initial guess were sampled from uniform distributions with varying bounds.
The bounds are $\pm 0.5 \, \sigma \, \si{\meter}$ for the translation components and $\pm 2.5 \, \sigma \si{\degree}$ for the heading, for noise scales $\sigma = \{0, 1, \dots, 4\}$.
To evaluate the impact of using our learned weight masks we run ICP once with all point prior weights set to one as a baseline and once with weights sampled from weight masks generated by the network.

\renewcommand{\arraystretch}{1.25}
\begin{table*}[t!]
\centering
\caption{Test sequence RMSE results for lateral/longitudinal translation, and heading. The standard deviations for the normal distribution used to sample the initial transformation are shown under Noise Scale. Results are computed from converged samples, with the percent of such samples shown in the percent converged (Conv.) column. The percent of converged results that are within $0.05 \, \si{\metre}$ translation and $1 \si{\degree}$ rotation of the groundtruth is shown in the percent accurate (Acc.) column.}
\begin{tabular}{||cc||c|c|c||c|c||c|c|c||c|c||}
\hline
\hline
\multicolumn{2}{||c||}{Noise Scale} & \multicolumn{5}{c||}{Unweighted RMSE} & \multicolumn{5}{c||}{Weighted RMSE (ours)} \\
\cline{3-12}
[\si{\meter}] & [\si{\degree}] & Long. [\si{\metre}] & Lat. [\si{\metre}] & Head. [\si{\degree}] & Conv. [\si{\percent}] & Acc. [\si{\percent}] & Long. [\si{\metre}] & Lat. [\si{\metre}] & Head. [\si{\degree}] & Conv. [\si{\percent}]  & Acc. [\si{\percent}]\\
\hline
\hline
0.0 & 0.0 & 0.135  &  0.095  &  0.252  &  99.79  &  13.27  & \textbf{ 0.079 } & \textbf{ 0.062 } & \textbf{ 0.147 } & \textbf{ 99.99 } & \textbf{ 37.89 }\\
\hline
0.5 & 2.5 & 0.140  &  0.097  &  0.285  &  99.63  &  11.51  & \textbf{ 0.087 } & \textbf{ 0.065 } & \textbf{ 0.179 } & \textbf{ 99.96 } & \textbf{ 32.74 }\\
\hline
1.0 & 5.0 & 0.142  &  0.097  &  0.294  &  98.13  &  11.56  & \textbf{ 0.088 } & \textbf{ 0.065 } & \textbf{ 0.182 } & \textbf{ 99.57 } & \textbf{ 32.47 }\\
\hline
1.5 & 7.5 & 0.145  &  0.098  &  0.319  &  89.95  &  11.58  & \textbf{ 0.093 } & \textbf{ 0.071 } & \textbf{ 0.210 } & \textbf{ 97.15 } & \textbf{ 32.57 }\\
\hline
2.0 & 10.0 & 0.176  &  0.124  &  0.634  &  73.52  &  11.63  & \textbf{ 0.113 } & \textbf{ 0.096 } & \textbf{ 0.343 } & \textbf{ 88.63 } & \textbf{ 32.86 }\\
\hline
\hline
\end{tabular}
\label{tbl:rmse_result}
\end{table*}

\subsection{Results}
Three representative test samples showing the qualitative effects of the learned masks are shown in Figure \ref{fig:example_masks}.
The masks learn to highlight structural features such as building walls and highly reflective roadside signs and posts, which particularly help constrain point-to-point ICP.
They also learn to largely ignore unusable points regardless of the strength of their return.
These ignored points include those resulting from radar artefacts, noise, and even other vehicles on the road.
Overall, the masked radar pointclouds are visually much closer to the lidar maps than unweighted ones.

Table \ref{tbl:rmse_result} presents the root mean squared error (RMSE) ICP results for each state component for the unweighted baseline and using our learned weights.
The results are presented for different noise scales, with the uniform distribution bounds shown in the left table column for the translation ($\si{\meter}$) and rotation ($\si{\degree}$) components.
We compute the results from converged samples, as non-converged results tend to correspond to outliers that would be easily ignored in a localization pipeline.
Convergence is determined by whether the norm of the final step size is below $0.001$, and the percent of converged samples compared to the entire test set is reported.
Finally, the table also reports the percent of `accurate' converged results.
We define accuracy here as ICP results that converged to within $0.05 \si{\metre}$ translation and $1 \si{\degree}$ rotation of the groundtruth.
These bounds were chosen as representative of a desired safety envelope for autonomous driving.
We note that these bounds are quite ambitious for stand-alone ICP results, as a full localization pipeline would typically include motion priors and other constraints.

As can be seen, our method improves the RMSE in every component and ICP convergence at every initialization noise scale.
We also show a large improvement in the number of ICP results that converge to within our desired accuracy envelope. Using our learned weights we are able to converge to the desired localization accuracy for deployable autonomous vehicles almost a third of the time across a variety of reasonable initial errors. This is compared to the baseline, which can accomplish the same task only a little over a tenth of the time.

\begin{figure}[t!]
    \centering
    \includegraphics[width=0.9\linewidth]{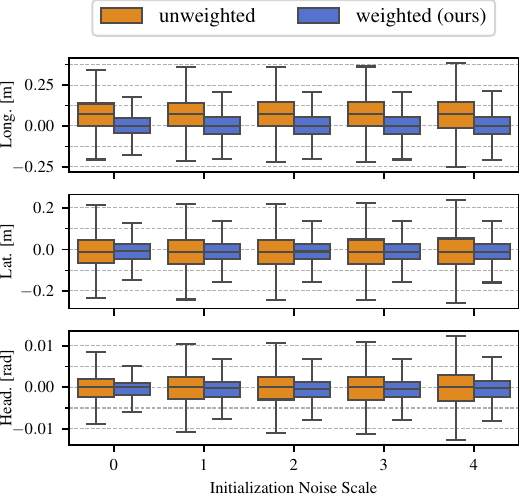}
    \caption{Box plot showing the distribution of errors between the baseline and our method in longitudinal translation, lateral translation, and heading.
    The results are presented for a changing noise scale, which is a factor applied to the boundaries of a uniform distribution from which initial guesses are sampled.
    Our method has a tighter error distribution in all components and across different initialization noise scales compared to the baseline.}
    \label{fig:result}
\end{figure}

Figure \ref{fig:result} shows the distribution of errors in each state component for the baseline and our method.
We observe that our method yields a significantly tighter distribution of errors across all noise scales.
We observe a bias in the longitudinal direction for the unweighted results.
This bias was similarly reported in \cite{are_we_ready_for}, and hypothesized to be a result of Doppler distortion effects\footnote[3]{Points from objects with a large relative velocity compared to the radar may be distorted in range due to the Doppler effect.}.
Interestingly, the use of the weighted masks seems to entirely remove this bias.
We hypothesize that the removal of the bias is due to distorted points failing to index into strongly weighted parts of the weight mask and thus no longer affecting ICP.
It is also likely that most Doppler-distorted points are found on vehicles, or that even undistorted points extracted from vehicles are detrimental to ICP.
Vehicles directly in front and behind the car, all moving in the same direction as the car, may disproportionately affect ICP in the longitudinal direction resulting in a bias.
As shown in Figure \ref{fig:example_masks}, we find that the weight masks learn to ignore vehicles on the road.
If the bias is due to other vehicles on the road, our masks ignore them effectively enough to entirely remove the bias.

\section{CONCLUSION}
    \label{sec:conclusion}
    This paper provides a novel method for improving radar-lidar localization by learning weights associated with radar pointclouds to be used in ICP.
This approach makes use of a localization pipeline with a heuristic point extractor and ICP.
Instead of replacing any functionality with an end-to-end network, we learn to generate a weight mask to be sampled by the heuristically extracted pointcloud.
The sampled weight is then used to control the value of radar points in a weighted ICP.
To train this network, we additionally introduce a publicly available novel implementation of a differentiable ICP algorithm.
Our approach ensures that the robustness and interpretability of an analytical pipeline is maintained, while reducing error and improving convergence in isolated radar-lidar ICP tests.
This makes radar-lidar localization more feasible to use in autonomous driving, providing greater safety and robustness in all weather conditions.

Future work will focus on integrating the learned weights into a full localization pipeline to see their impact in the presence of informed priors on ICP. Finally, it is of interest to evaluate the generalizability of this network to new maps, or to different radar sensors on the same map.

\renewcommand*{\bibfont}{\footnotesize}
\printbibliography

@article{survey_lidar_loc,
  title={{A Survey on Global LiDAR Localization: Challenges, Advances and Open Problems}},
  author={Yin, Huan and Xu, Xuecheng and Lu, Sha and Chen, Xieyuanli and Xiong, Rong and Shen, Shaojie and Stachniss, Cyrill and Wang, Yue},
  journal={Int. J. Comput. Vis.},
  pages={1--33},
  year={2024},
  publisher={Springer}
}

@article{are_we_ready_for,
  title={{Are We Ready for Radar to Replace Lidar in All-Weather Mapping and Localization?}},
  author={Burnett, Keenan and Wu, Yuchen and Yoon, David J and Schoellig, Angela P and Barfoot, Timothy D},
  journal={IEEE Robot. Automat. Lett.},
  volume={7},
  number={4},
  pages={10328--10335},
  year={2022},
  publisher={IEEE}
}

@article{yoneda2019automated,
  title={{Automated driving recognition technologies for adverse weather conditions}},
  author={Yoneda, Keisuke and Suganuma, Naoki and Yanase, Ryo and Aldibaja, Mohammad},
  journal={IATSS Research},
  volume={43},
  number={4},
  pages={253--262},
  year={2019},
  publisher={Elsevier}
}

@inbook{Courcelle_Baril_Pomerleau_Laconte_2022,
  author = {Courcelle, Clément and Baril, Dominic and Pomerleau, François and Laconte, Johann},
  editor = {Abut, Huseyin and Schmidt, Gerhard and Takeda, Kazuya and Lambert, Jacob and Hansen, John H.L.},
  title = {{On the Importance of Quantifying Visibility for Autonomous Vehicles under Extreme Precipitation}},
  booktitle = {Towards Human-Vehicle Harmonization},
  volume = {3},
  year = {2023},
  publisher = {De Gruyter},
  pages = {239--250},
  doi = {doi:10.1515/9783110981223-018},
  isbn = {9783110981223},
  project = {snow}
}

@inproceedings{Park_Kim_Kim_2019,
  title={{Radar Localization and Mapping for Indoor Disaster Environments via Multi-modal Registration to Prior LiDAR Map}},
  author={Park, Yeong Sang and Kim, Joowan and Kim, Ayoung},
  booktitle={2019 IEEE/RSJ Int. Conf. Intell. Robots Syst.},
  pages={1307--1314},
  year={2019},
  organization={IEEE}
}

@INPROCEEDINGS{weather_sensor_impact,
  author={Sezgin, Fatih and Vriesman, Daniel and Steinhauser, Dagmar and Lugner, Robert and Brandmeier, Thomas},
  booktitle={2023 IEEE Intell. Vehicles Symp. (IV)}, 
  title={{Safe Autonomous Driving in Adverse Weather: Sensor Evaluation and Performance Monitoring}}, 
  year={2023},
  volume={},
  number={},
  pages={1-6},
  doi={10.1109/IV55152.2023.10186596}
}

@book{radar_nav_book,
  title={{Robotic Navigation and Mapping with Radar}},
  author={Adams, Martin and Adams, Martin David},
  year={2012},
  publisher={Artech House}
}

@inproceedings{Weston_Jones_Posner_2020,
  title={{There and Back Again: Learning to Simulate Radar Data for Real-world Applications}},
  author={Weston, Rob and Jones, Oiwi Parker and Posner, Ingmar},
  booktitle={2021 IEEE Int. Conf. Robot. Automat.},
  pages={12809--12816},
  year={2021},
  organization={IEEE}
}

@article{RaLL,
  title={{RaLL: End-to-end Radar Localization on Lidar Map Using Differentiable Measurement Model}},
  author={Yin, Huan and Chen, Runjian and Wang, Yue and Xiong, Rong},
  journal={IEEE Trans. Intell. Transp. Syst.},
  volume={23},
  number={7},
  pages={6737--6750},
  year={2021},
  publisher={IEEE}
}

@inproceedings{RoLM,
  title={{RoLM: Radar on LiDAR Map Localization}},
  author={Ma, Yukai and Zhao, Xiangrui and Li, Han and Gu, Yaqing and Lang, Xiaolei and Liu, Yong},
  booktitle={2023 IEEE Int. Conf. Robot. Automat.},
  pages={3976--3982},
  year={2023},
  organization={IEEE}
}

@inproceedings{icp_ref,
  title={{Method for registration of 3-D shapes}},
  author={Besl, Paul J and McKay, Neil D},
  booktitle={Sensor Fusion IV: Control Paradigms and Data Structures},
  volume={1611},
  pages={586--606},
  year={1992},
  organization={Spie}
}

@article{radar_survey,
  title={{A New Wave in Robotics: Survey on Recent mmWave Radar Applications in Robotics}},
  author={Harlow, Kyle and Jang, Hyesu and Barfoot, Timothy D and Kim, Ayoung and Heckman, Christoffer},
  journal={IEEE Trans. Robot.},
  year={2024},
  publisher={IEEE}
}

@ARTICLE{radar_survey_2,
  author={Abu-Alrub, Nader J. and Rawashdeh, Nathir A.},
  journal={IEEE Trans. Intell. Vehicles}, 
  title={{Radar Odometry for Autonomous Ground Vehicles: A Survey of Methods and Datasets}}, 
  year={2024},
  volume={9},
  number={3},
  pages={4275-4291},
  doi={10.1109/TIV.2023.3340513}
}

@article{venon2022millimeter,
  title={{Millimeter Wave FMCW Radars for Perception, Recognition and Localization in Automotive Applications: A Survey}},
  author={Venon, Arthur and Dupuis, Yohan and Vasseur, Pascal and Merriaux, Pierre},
  journal={IEEE Trans. Intell. Vehicles},
  volume={7},
  number={3},
  pages={533--555},
  year={2022},
  publisher={IEEE}
}

@inproceedings{Cen_Newman_2018,
  title={{Precise Ego-motion Estimation with Millimeter-wave Radar Under Diverse and Challenging Conditions}},
  author={Cen, Sarah H and Newman, Paul},
  booktitle={2018 IEEE Int. Conf. Robot. Automat.},
  pages={6045--6052},
  year={2018},
  organization={IEEE}
}

@inproceedings{masking_by_moving,
  author = {Dan Barnes and Rob Weston and Ingmar Posner},
  title = {{Masking by Moving: Learning Distraction-Free Radar Odometry from Pose Information}},
  booktitle = {{C}onference on {R}obot {L}earning ({CoRL})},
  url = {https://arxiv.org/pdf/1909.03752},
  year = {2019}
}

@inproceedings{under_the_radar,
  author = {Dan Barnes and Ingmar Posner},
  title = {{Under the Radar: Learning to Predict Robust Keypoints for Odometry Estimation and Metric Localisation in Radar}},
  booktitle = {IEEE Int. Conf. Robot. Automat.},
  year = {2020}
}

@INPROCEEDINGS{hero_paper,
    title={{Radar Odometry Combining Probabilistic Estimation and Unsupervised Feature Learning}},
    author={Burnett, Keenan and Yoon, David J and Schoellig, Angela P and Barfoot, Timothy D},
    booktitle={Robot.: Sci. Syst.},
    year={2021}
}

@inproceedings{Jose_Adams_2004,
  title={{Relative RADAR Cross Section Based Feature
Identification with Millimetre Wave RADAR for
Outdoor SLAM}},
  author={Jose, Ebi and Adams, Martin David},
  booktitle={2004 IEEE/RSJ Int. Conf. Intell. Robots Syst.},
  volume={1},
  pages={425--430},
  year={2004},
  organization={IEEE}
}

@inproceedings{Rouveure_Monod_Faure_2009,
  title={{High Resolution Mapping of the Environment with a Ground-based Radar Imager}},
  author={Rouveure, R and Monod, MO and Faure, P},
  booktitle={2009 Int. Radar Conf. ``Surveillance for a Safer World'' (RADAR 2009)},
  pages={1--6},
  year={2009},
  organization={IEEE}
}

@inproceedings{Checchin_Gerossier_Blanc_Chapuis_Trassoudaine_2010,
  title={{Radar Scan Matching SLAM Using the Fourier-mellin Transform}},
  author={Checchin, Paul and G{\'e}rossier, Franck and Blanc, Christophe and Chapuis, Roland and Trassoudaine, Laurent},
  booktitle={Field and Service Robotics: Results of the 7th International Conference},
  pages={151--161},
  year={2010},
  organization={Springer}
}

@inproceedings{Schuster_Keller_Rapp_Haueis_Curio_2016,
  title={{Landmark based radar SLAM using graph optimization}},
  author={Schuster, Frank and Keller, Christoph Gustav and Rapp, Matthias and Haueis, Martin and Curio, Crist{\'o}bal},
  booktitle={2016 IEEE 19th Int. Conf. Intell. Transp. Syst.},
  pages={2559--2564},
  year={2016},
  organization={IEEE}
}

@inproceedings{Ward_Folkesson_2016,
  title={{Vehicle Localization with Low Cost Radar Sensors}},
  author={Ward, Erik and Folkesson, John},
  booktitle={2016 IEEE Intell. Vehicles Symp. (IV)},
  pages={864--870},
  year={2016},
  organization={IEEE}
}

@article{cfar,
  title={{Radar CFAR Thresholding in Clutter and Multiple Target Situations}},
  author={Rohling, Hermann},
  journal={IEEE Trans. Aerosp. Electron. Syst.},
  number={4},
  pages={608--621},
  year={1983},
  publisher={IEEE}
}

@article{bfar,
  title={{BFAR-bounded False Alarm Rate Detector for Improved Radar Odometry Estimation}},
  author={Alhashimi, Anas and Adolfsson, Daniel and Magnusson, Martin and Andreasson, Henrik and Lilienthal, Achim J},
  journal={arXiv preprint {\tt arXiv:2109.09669}},
  year={2021}
}

@inproceedings{radar_on_lidar,
  title={{Radar-on-lidar: Metric Radar Localization on Prior Lidar Maps}},
  author={Yin, Huan and Wang, Yue and Tang, Li and Xiong, Rong},
  booktitle={2020 IEEE Int. Conf. Real-time Comput. Robot.},
  pages={1--7},
  year={2020},
  organization={IEEE}
}

@article{Yin_Xu_Wang_Xiong_2021,
  title={{Radar-to-lidar: Heterogeneous Place Recognition via Joint Learning}},
  author={Yin, Huan and Xu, Xuecheng and Wang, Yue and Xiong, Rong},
  journal={Frontiers in Robotics and AI},
  volume={8},
  pages={661199},
  year={2021},
  publisher={Frontiers Media SA}
}

@inproceedings{unet,
  title={{U-Net: Convolutional Networks for Biomedical Image Segmentation}},
  author={Ronneberger, Olaf and Fischer, Philipp and Brox, Thomas},
  booktitle={Medical Image Computing and Computer-Assisted Intervention (MICCAI)},
  pages={234--241},
  year={2015},
  organization={Springer}
}

@article{dropout,
  title={{Dropout: A Simple Way to Prevent Neural Networks from Overfitting}},
  author={Srivastava, Nitish and Hinton, Geoffrey and Krizhevsky, Alex and Sutskever, Ilya and Salakhutdinov, Ruslan},
  journal={J. Mach. Learn. Res.},
  volume={15},
  number={1},
  pages={1929--1958},
  year={2014},
  publisher={JMLR. org}
}

@article{pytorch,
  title={{PyTorch: An Imperative Style, High-performance Deep Learning Library}},
  author={Paszke, Adam and Gross, Sam and Massa, Francisco and Lerer, Adam and Bradbury, James and Chanan, Gregory and Killeen, Trevor and Lin, Zeming and Gimelshein, Natalia and Antiga, Luca and others},
  journal={Advances Neural Inf. Process. Syst.},
  volume={32},
  year={2019}
}

@book{barfoot2017state,
  title={{State Estimation for Robotics}},
  author={Barfoot, T.D.},
  isbn={9781107159396},
  lccn={2017010237},
  year={2017},
  publisher={Cambridge, UK: Cambridge University Press}
}

@inproceedings{pypose,
  title={{PyPose: A Library for Robot Learning with Physics-based Optimization}},
  author={Wang, Chen and Gao, Dasong and Xu, Kuan and Geng, Junyi and Hu, Yaoyu and Qiu, Yuheng and Li, Bowen and Yang, Fan and Moon, Brady and Pandey, Abhinav and others},
  booktitle={Proc. IEEE/CVF Conf. Comput. Vis. Pattern Recognit.},
  pages={22024--22034},
  year={2023}
}

@article{grad_slam,
  title={{Grad-SLAM: Explaining Convolutional Autoencoders’ Latent Space of Satellite Image Time Series}},
  author={Di Martino, Thomas and Guinvarc’h, R{\'e}gis and Thirion-Lefevre, Laetitia and Colin, {\'E}lise},
  journal={IEEE Geosci. Remote Sens.},
  year={2023},
  publisher={IEEE}
}

@article{binbin_paper,
  title={{Deep Probabilistic Feature-metric Tracking}},
  author={Xu, Binbin and Davison, Andrew J and Leutenegger, Stefan},
  journal={IEEE Robot. Automat. Lett.},
  volume={6},
  number={1},
  pages={223--230},
  year={2020},
  publisher={IEEE}
}

@inproceedings{gumbel_softmax,
  title={{Categorical Reparameterization with Gumbel-Softmax}},
  author={Jang, Eric and Gu, Shixiang and Poole, Ben},
  booktitle={Int. Conf. Learn. Representations},
  year={2016}
}

@book{hartley2003multiple,
  title={{Multiple View Geometry in Computer Vision}},
  author={Hartley, Richard and Zisserman, Andrew},
  year={2003},
  publisher={Cambridge University Press}
}

@article{VTR,
  title={{Visual Teach and Repeat for Long-range Rover Autonomy}},
  author={Furgale, Paul and Barfoot, Timothy D},
  journal={J. Field Robot.},
  volume={27},
  number={5},
  pages={534--560},
  year={2010},
  publisher={Wiley Online Library}
}

@article{boreas,
author = {Keenan Burnett and David J Yoon and Yuchen Wu and Andrew Z Li and Haowei Zhang and Shichen Lu and Jingxing Qian and Wei-Kang Tseng and Andrew Lambert and Keith YK Leung and Angela P Schoellig and Timothy D Barfoot},
title ={{Boreas: A Multi-season Autonomous Driving Dataset}},
journal = {Int. J. Robot. Res.},
volume = {42},
number = {1-2},
pages = {33-42},
year = {2023},
doi = {10.1177/02783649231160195}
}

@InProceedings{adam,
  author    = {Kingma, Diederik and Ba, Jimmy},
  booktitle = {Int. Conf. Learn. Representations},
  title     = {{Adam: A Method for Stochastic Optimization}},
  year      = {2015}
}

@INPROCEEDINGS{robust_func,
  author={Babin, Philippe and Giguère, Philippe and Pomerleau, François},
  booktitle={2019 Int. Conf. Robot. Automat.}, 
  title={{Analysis of Robust Functions for Registration Algorithms}}, 
  year={2019},
  volume={},
  number={},
  pages={1451-1457},
  doi={10.1109/ICRA.2019.8793791}
}

@article{reid2019localization,
  title={{Localization Requirements for Autonomous Vehicles}},
  author={Reid, Tyler GR and Houts, Sarah E and Cammarata, Robert and Mills, Graham and Agarwal, Siddharth and Vora, Ankit and Pandey, Gaurav},
  journal={arXiv preprint {\tt arXiv:1906.01061}},
  year={2019}
}

\end{document}